\documentclass[journal]{IEEEtran}

\usepackage{cite}           % 导入引用的包
\usepackage{balance}        % 在ref的时候让左右栏平衡
\usepackage{amsmath}        % 导入数学公式包
\usepackage{graphicx}
\usepackage{float}
\usepackage{stfloats}
\usepackage{makecell}       % 使用此包可进行表格内换行

\begin{document}

% 文章标题
\title{AuE-IPA: An AU Engagement Based Infant Pain Assessment Method}

% 作者姓名和职务介绍
\author{Mingze~Sun, Haoxiang~Wang, Wei~Yao, Jiawang~Liu
\thanks{M. Sun et al. is with the School of Computer Science and Engineering,
South China University of Technology, China; e-mail: 1836828080@qq.com.}}

% 文档题头
\markboth{Journal of \LaTeX\ Class Files,~Vol.~14, No.~8, August~2015}%
{Shell \MakeLowercase{\textit{et al.}}: Bare Demo of IEEEtran.cls for IEEE Journals}

\maketitle

% 摘要
\begin{abstract}
Recent studies have found that pain in infancy has a significant impact on infant development, including psychological problems, possible brain injury, and pain sensitivity in adulthood. However, due to the lack of specialists and the fact that infants are unable to express verbally their experience of pain, it is difficult to assess infant pain. Most existing infant pain assessment systems directly apply adult methods to infants ignoring the differences between infant expressions and adult expressions. Meanwhile, as the study of facial action coding system continues to advance, the use of action units (AUs) opens up new possibilities for expression recognition and pain assessment. In this paper, a novel AuE-IPA method is proposed for assessing infant pain by leveraging different engagement levels of AUs. First, different engagement levels of AUs in infant pain are revealed, by analyzing the class activation map of an end-to-end pain assessment model. The intensities of top-engaged AUs are then used in a regression model for achieving automatic infant pain assessment. The model proposed is trained and experimented on YouTube Immunization dataset, YouTube Blood Test dataset, and iCOPEVid dataset. The experimental results show that our AuE-IPA method is more applicable to infants and possesses stronger generalization ability than end-to-end assessment model and the classic PSPI metric.
\end{abstract}

% 关键词

% Introduction
\section{Introduction}
% 疼痛检测的必要性
\IEEEPARstart{W}{ith} the rapid development of artificial intelligence, people make great progress in human-computer interaction technologies and affective computing. As a major part of affective computing, facial expression recognition receives much attention from scholars and scientists\cite{9697318}. Facial expressions can be considered as a valid indicator of a person’s degree of pain and pain assessment is of great benefit to healthcare. The International Association for the Study of Pain (IASP) recently revised pain definition as ``An unpleasant sensory and emotional experience associated with, or resembling that associated with, actual or potential tissue damage\cite{raja2020revised}.'' Pain is also the most common clinical manifestation and many patients seek medical suggestions for pain.

% 婴儿疼痛检测的必要性
It is therefore essential to assess pain intensity and apply effective pain management strategies promptly. This is particular true for infants. Pain in infancy affects mental health later in life and impact pain reactivity\cite{brummelte2012procedural}. Limited by infants' faculty of expressing themselves, infants can't articulate their sensations effectively. Additionally, owing to the observer bias, caregivers and non-professionals have trouble assessing pain accurately. So it is necessary to achieve an automatic pain assessment model for infants.
\begin{figure}[t]
    \centering
    \includegraphics[width=0.48\textwidth]{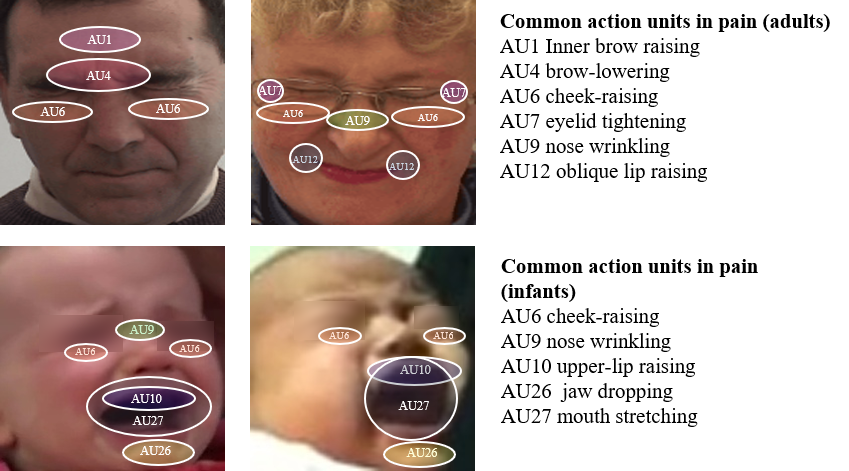}
    \caption{The comparisons of infants and adults with different expressions and action units in pain. The adult images are from UNBC shoulder dataset\cite{5771462}, and the infants images are from YouTube Immunization dataset\cite{harrison2014too}.}
    \label{fig:AdultNeonatalFig}
\end{figure}

% 先说现在方法存在的问题(婴儿适用性，范化性问题)
% 再说使用面部表情进行识别的原因
There have been a lot of researches on automatic pain assessment\cite{8120021}. But most of the present approaches neglect the differences between infant expressions and adult expressions. Due to acquired educational and social factors, the facial expressions of infant pain differ greatly from those of adults, as shown in Fig \ref{fig:AdultNeonatalFig}. Infants have a stronger response to pain and more mouth actions. Furthermore, many researches were experimented on a single public dataset or non-public private data which makes these models difficult to be applied in wild scenarios. Approaches such as those in \cite{brahnam2020neonatal, BRAHNAM20071242, nanni2010local, martinez2020application, 8851879} are performed on a single dataset. While the approaches of [7,8] were performed on a private self-collected dataset. Since the datasets are relatively small in size for deep neural network models, there is often a problem of over-fitting and insufficient generalization. Some recognition\cite{8120021} approaches use a variety of modal information, including facial expressions, infant cries, body movement, vital signs, etc. However, in realistic scenarios, such as homes and hospital facilities with basic or rudimentary settings, it is difficult to collect accurate multimodal data. Consequently, a more comprehensive, applicable, and feasible method for infant pain assessment is needed.

% FACS 的定义以及使用FACS 的好处
The Facial Action Coding System (FACS) is a system for describing facial muscle movements\cite{cohn2007observer}. It decomposes facial expressions into facial muscle movements, i.e. action units (AU). FACS is widely used for expression and micro-expression recognition. There are also many practices of using AUs for pain recognition. The Prkachin and Solomon pain intensity (PSPI) metric is derived by counting the AU with a high probability of occurrence in the pain expressions of patients suffering from shoulder pain. Grunau et al. \cite{GRUNAU1987395} proposed a Neonatal Facial Coding System (NFCS) by analyzing the facial expressions of infants. Facial expressions in pain can be accurately judged using FACS and AU data is effective in eliminating individual-related factors to address the problem of over-fitting due to the small size of datasets. AU based pain assessment has became feasible with the recent development of AU detection models and tools\cite{7284869, 8373812, 8880512}. 

% 使用 FACS 方法的不足之处
However, current pain assessment approaches using FACS are inadequate for infants. The PSPI pain assessment metric derived from adult experiments is not well suited for infants. With regard to the action units used in NFCS, they are not mainstream and are not supported by popular AU detection models. In addition, many present assessment methods using FACS do not give careful consideration on the selection of AUs. They only select specific AUs based on previous experience and neglect different engagement levels of AUs in pain expression.

% 介绍我们的工作和贡献
In this paper, we propose a novel AU Engagement based Infant Pain Assessment(AuE-IPA) method. We conduct the pioneering work of exploring different engagement levels of AUs in infant pain, by analyzing the Gradient Class Activation Map(Grad-CAM)\cite{8237336} of an end-to-end pain assessment model. Based on engagement levels, the most crucial AUs are selected for assessing infant pain using a regression model. The main contributions of this paper can be summarized as follows:
\begin{itemize}
    \item To the best of our knowledge, this is the first work of revealing the most important AUs for infant pain assessment, based on their engagement levels learned from an end-to-end deep neural network.
    \item A regression model is presented as a new infant pain assessment metric. The metric utilizes both intensities and importance of top-engaged AUs.
    \item The novelty of our AuE-IPA method has been proved by sufficient experiments in two aspects: the AUs selected play important role in infant pain assessment; the new infant pain assessment metric is more applicable to infants and possesses stronger generalization ability than end-to-end assessment model and the classic PSPI metric.
\end{itemize}
% 介绍全文行文结构
Organization: Section II describes related works to this paper. Section III proposes the end-to-end assessment model and the engagement levels based method. Section IV introduces the datasets used in the experiment and their pre-processing steps. Part V presents the experimental results, and in Part VI we summarizes the overall work.

% Related Works
\section{Related Works}
% 介绍成人疼痛评估方法, 和成人疼痛
\subsection{Adult Pain Intensity Assessment}
According to the FACS study\cite{cohn2007observer}, facial expressions can be broken down into different facial action units. Pain expressions are also closely associated with multiple facial action units. To understand human facial expressions in pain, researchers at McMaster University and University of Northern British Columbia experimented with people suffering from shoulder pain to record their expressions and facial action units\cite{prkachin1992consistency, PRKACHIN2008267, 5771462}.

In the study of shoulder pain in adults, by analyzing the facial expressions and counting the frequency of different facial action units, the researchers obtained 11 AUs that are closely related to pain\cite{prkachin1992consistency}. The 11 AUs are brow-lowering (AU4), cheek-raising (AU6), eyelid tightening (AU7), nose wrinkling (AU9), upper-lip raising (AU10), oblique lip raising (AU12), horizontal lip stretch (AU20), lips parting (AU25), jaw dropping (AU26), mouth stretching (AU27) and eye closure (AU43). In a follow-up study, Prkachin et al.\cite{PRKACHIN2008267} suggested that the following core AU carries most of the pain information and used them to determine the Prkachin and Solomon pain intensity (PSPI) metric:
% PSPI 公式
\begin{equation} 
    \label{equ:PSPIEqu}
    \begin{split}
    Pain = &AU4 + max(AU6, AU7) + \\
    &max(AU9, AU10) + AU43
    \end{split}
\end{equation}

As described in the Equation \ref{equ:PSPIEqu}, the PSPI uses the sum of intensities of the 6 core AUs to describe pain. PSPI metric illustrates the association between pain and facial expressions and it has been widely used for pain annotation in adults. For example, the BioVid Heat Pain Database records the experimenter's face and movement units and uses PSPI to annotate frame-level pain intensity\cite{7423704}. There are also many methods that use the PSPI indicator as experimental indicator. Bargshady et al.\cite{BARGSHADY2020101954,BARGSHADY2020106805,BARGSHADY2020113305} extracted portrait features by fine-tuning the pre-trained model (VGG-Net). They performed principal component analysis (PCA) to reduce the dimension of the extracted features and transformed the color space. They achieved good results on the UNBC dataset. Huang et al.\cite{huang_hybnet_2022} combined multiple information including 2D features, facial coordinate features, and sequence features to regress the PSPI metric.

However, there is a clear difference between infants' expressions of pain and adults' expressions of pain. Adults make pain expressions not only as a physiological motor response but also influenced by social and cultural influences. The core AUs defined by PSPI (brow-lowering, cheek-raising, eye lid tightening, nose wrinkling, upper-lip raising, eye closure) are related to the upper part of the face, while infants in pain tend to cry out and take more action units related to lower part of the face. In other words, adults express pain more subtly and infants express pain more directly. Studies have also shown that infants' facial pain expressions are more consistent than those of adults\cite{2001-05101-009}. Therefore, PSPI does not describe infant pain well, and we need a better AU-based metric for infants.

% 介绍婴儿疼痛并说明不足之处
\subsection{Infant Pain \& Infant Pain Intensity Assessment}
Due to the lack of professional medical facilities in many scenarios, pain detection using facial expressions is still the simplest and most effective method. Recently, automatic pain detection system for infants has also gained great attention. 

Brahnam et al. \cite{BRAHNAM20071242} used grayscale features of the infant's face on the COPE dataset. They performed PCA reduction and applied the reduced features to classification. While Nanni et al. \cite{nanni2010local} used local binary pattern (LBP) as a feature extractor, then extracted features from the infant's face and classified using support vector machines (SVM). Martínez et al. \cite{martinez2020application} used the Radon Barcodes (RBC) feature extractor, which is mostly used in medical imaging, and cropped the regions related to the baby's AUs. They extracted features from these cropped regions and applied them to classification.
% 整体结构图
\begin{figure*}[hb] 
    \centering
    \includegraphics[width=\textwidth]{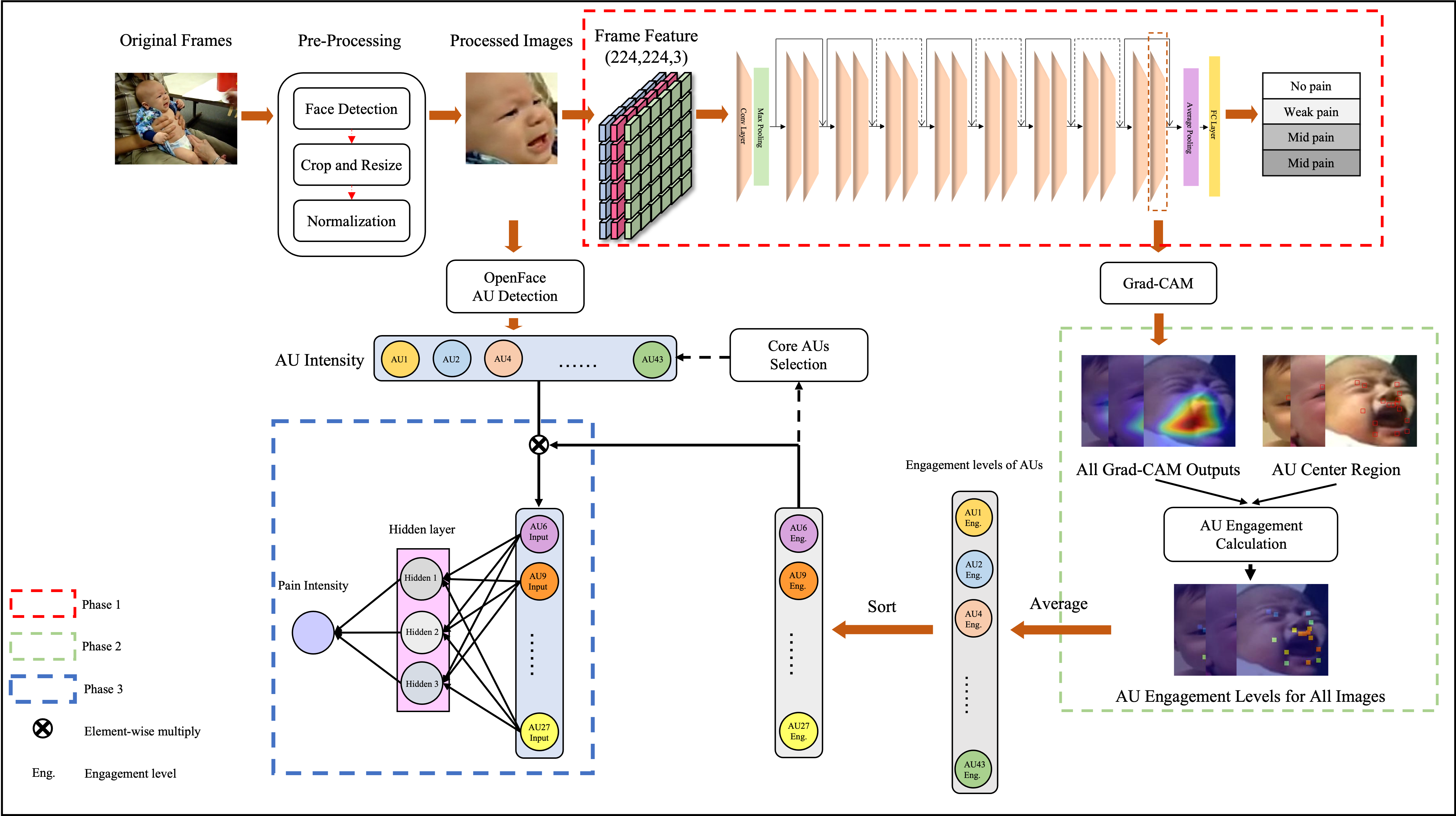}
    \caption{\textbf{Overall structure of AU Engagement Based Infant Pain Assessment Method (AUE-IPA).} The method is mainly composed of three phases: 1.The end-to-end pain assessment model uses the infant's face as input to evaluate the pain intensity for learning the infant's facial expressions in pain (red dashed box).; 2. The different engagement levels of AUs are computed by analyzing the Grad-CAM\cite{8237336} output of this end-to-end deep network (green dashed box).; 3. The most crucial AUs are selected according to the engagement levels and fed into pain intensity regression model (blue dashed box).}
    \label{fig:StructureFig}
\end{figure*}

Neural network-based models likewise have a lot of practice. Zamzmi et al.\cite{8851879} proposed Neonatal Convolutional Neural Network (N-CNN) by combining convolutional neural networks with convolutional kernels of different sizes. They used N-CNN for feature extraction and classification of baby pictures and validated it on the COPE dataset. Liu et al.\cite{9450899}, on the other hand, used a modified AlexNet with infant pictures as input and experimented it on a larger, self-collected infant pain dataset. The experimental results demonstrate that the model outperforms LBP and Histogram of Oriented Gradients (HOG).

Some researchers have also used the FACS method for infant pain detection. Based on prior knowledge of the classic FACS, sikka et al.\cite{sikka2015automated} regressed pain levels using the intensity of pain-related AUs, however their experiments were based on an unpublished dataset collected in hospitals. Considering the differences between infants and adults, Grunau et al. analyzed facial pain expressions during heel-lance and blood collection on infants and proposed the Neonatal Facial Coding System(NFCS)\cite{GRUNAU1987395}. They calculated the frequency of different facial action units in pain and found that brow bulge, eye squeeze, nasolabial furrow, and stretch open mouth were more frequent. 

Previous pain assessment models neglect baby's characteristic and simply apply adult methods to infants. These methods are performed on single datasets that are small in size, e.g., the COPE dataset\cite{brahnam2005svm} has only 200 photos of 26 children. Neural network models tend to have over-fitting problems and poor generalization ability. 

Besides, the action units used in NFCS are not mainstream and not supported by popular AU detection models. Present assessment methods using FACS only select specific AUs based on previous experience and neglect different engagement levels of AUs in pain expression. In contrast, the AuE-IPA method we proposed aims to explore different engagement levels of AUs in infant pain. We select the most crucial AUs based on engagement levels for assessing infant pain.
% AU Center 图和 AU 效果图
\begin{figure*}[hb]     
    \centering
    \includegraphics[scale=0.55]{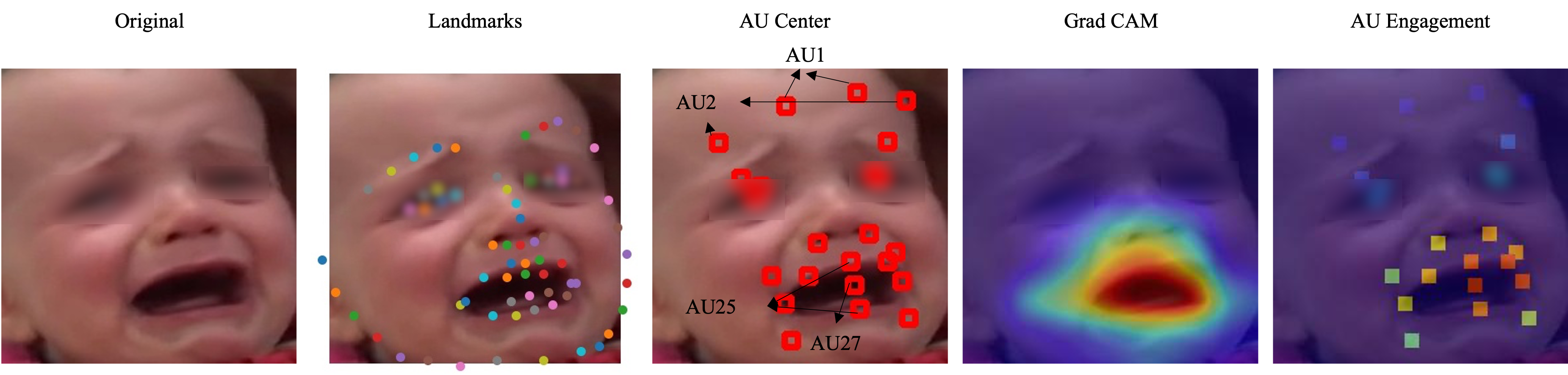}
    \caption{An example of calculating different engagement levels of AUs in infant pain. AU centers are derived from landmarks according to the center rules and AU-related regions are divided with size of $10 \times 10$. By calculating pixel values of the AU-related region on CAM outputs, the engagement levels are obtained.}
    \label{fig:AUCenterFig}
\end{figure*}

% Method
\section{Method}
% 介绍整体流程
% TODO 这里欠缺一个算法流程图
\subsection{The Overall Framework}
The structure of the approach is shown in Fig \ref{fig:StructureFig}. In the first phase, we fine-tune the pre-trained ResNet\cite{7780459} model to assess infant pain. Residual structure is proposed in ResNet model for addressing problems such as accuracy degradation and gradient explosion. The Grad-CAM layer is added to the transferred end-to-end model to analyse what the neural networks learn. Subsequently, we define the AU centers rule for the baby's face and use CAM outputs of the end-to-end model to calculate AUs engagement levels. In the last part, we conduct experiments using intensities of different AU depending on the levels of engagement. We finally reveal the core AUs in infant pain and propose the regression model using both intensities and importance of core AUs.

% 介绍端到端的面部疼痛识别模型, Grad-CAM 的添加
\subsection{End-to-End Pain Assessment Model}
First, we extract the visual frames from the raw data detailed in Chapter \ref{datasets}. Since two datasets (YouTube Immunization database \& YouTube Blood Test database) are derived from videos uploaded by YouTube users, infant images in videos have different orientations, locations and scales. So we detect and crop the child's faces using Dual Shot Face Detector (DSFD) \cite{li2018dsfd} on the selected frames. DSFD is an effective and easy-to-use face detector that was pre-trained using the WIDER face dataset\cite{yang2016wider}. We resized the cropped baby faces to $ 224 \times 224 \times 3 $ pixels because this representation is the most commonly used and compatible with most deep neural networks.

The ResNet model was proposed by He et al. for solving the degradation problem of deep neural network models with deeper layers by introducing the residual structure\cite{7780459}, and this model works well on numerous datasets in many domains. ResNet (ResNet18 is used in our method) consists of one convolutional layer, one global pooling layer, four residual convolutional layers, one average pooling layer, and one fully connected layer from top to bottom. Each residual convolutional layer contains two basic-blocks, and each basic-block is composed of two convolutional layers with residual structure. Thus, overall, ResNet18 consists of 17 convolutional layers and one fully-connected layer.

Transfer learning for image classification is proved to be effective in many fields. The ResNet18 model trained on ImageNet\cite{5206848} has learned a wealth of knowledge. In order to have a deep understanding of the infant's images and pain, we fine-tune the parameters of the pre-trained model by conducting pain classification on the YouTube Immunization dataset. Since the immunization dataset differs significantly from the ImageNet used for pre-training, we modify all parameters of the model and replace the fully connected layer with a new fully connected layer. The size of the replaced layer is changed to 4 which is the number of pain classes in YouTube Immunization dataset. We saved the parameters of the model that performed the best on the dataset.

% TODO 写到 CAM 层的输出和补全像素以及公式
In order to know the importance of different infant facial regions in pain classification, we use Grad-CAM\cite{8237336} to analyze the knowledge learned by the neural network. Grad-CAM uses the gradients of the final convolutional layer to produce a localization map highlighting important regions in the image. In this work, Grad-CAM tells us which facial regions the ResNet18 model focuses on for pain assessment. We compute the CAM output of the last convolutional layer of the model, which is in the last basic-block, because it is the closest to the classification target.

% AU 参与度计算方法
% AU Center 可以尝试举几个例子在图中解释
\subsection{Engagement Levels of AUs}
\label{sec:Engage}
After obtaining the CAM output of the model, we use the AU Centers method to further identify the engagement levels of different AUs in infant pain. As illustrated in previous literature and practice, different AU are highly correlated with corresponding sub-areas of the face\cite{li2017eac, martinez2020application}. The AU center, the center of each AU-related region, is determined by the position of the facial muscle group which controls corresponding AU. 
% AU Center 定义表
% AU 的名字必须得统一
\begin{table}[ht]
\small
\centering
\caption{Pain-related AU centers rules for infants}
\label{tab:AURuleTab}
\setlength{\tabcolsep}{6pt}
\renewcommand{\arraystretch}{1.2}
\begin{tabular}{|c |c| c|}
\hline
\textbf{AU Index}& \textbf{AU Name}& \textbf{AU Center}\\
\hline
1&Inner Brow Raiser&1/2 scale above inner brow\\
2&Outer Brow Raiser&1/3 scale above outer brow\\
4&Brow Lowerer&1/3 scale below brow center\\
6&Check Raiser&1 scale below eye bottom\\
7&Lid Tightener&Eye center\\
9&Nose Wrinkler&Ala of nose\\
10&Upper Lip Raiser&Upper lip center\\
12&Lip Corner Puller&Lip corner\\
20&Lip Strecher&1 scale below lip corner\\
25&Lips Part&Lip center\\
27&Mouth stretch&Mouth center\\
43&Eye-closure&Eye center\\
\hline
\end{tabular}
\end{table}

We take AUs listed in Table \ref{tab:AURuleTab} into consideration, as those are present in datasets and very relevant to pain\cite{cohn2007observer, 5771462}. The AUs center of infant are pre-defined based on the rules derived from specific facial muscle and definitions in \cite{li2017eac, fan2020facial, shao2021jaa}. AU-related regions are divided using the AU centers, as the central points with size of $10 \times 10$. Take AU1 in Fig \ref{fig:AUCenterFig} for example, the center locates at 1/2 scaled distance above inner brow. The scaled distance is defined as a reference for individual differences by calculating the distance between two eyes. The AU1-related region is derived from this rule and landmarks as shown in Fig \ref{fig:AUCenterFig} third column.

Based on the CAM output, the engagement levels of different AUs can be evaluated by averaging pixel values in the regions around each AU center. Only correctly classified images are selected for consideration and engagement level for each AU is calculated following Equation \ref{equ:AUEquation}. In this equation, $ N $ represents the set of correctly classified infant pictures and $ S $ is the area of the AU-related region which equals 100. $ CA $ represents the pixel value in CAM output and $ x, y $ are the coordinates of corresponding AU related region. The engagement levels are shown in Fig \ref{fig:AUEngagement}, and details of the experiments are described in Section \ref{sec:Engage}.
\begin{equation} 
    \label{equ:AUEquation}
    \begin{split}
    AU_{n} E &= \frac{\sum_{i=1}^{N} \sum_{x=C_{i}^{AU_{n}X_{l}}}^{ C_{i}^{AU_{n}X_{r}}} \sum_{y=C_{i}^{AU_{n}Y_{t}}}^{ C_{i}^{AU_{n}Y_{b}}} CA_i(x,y)}{N * S}\\
    \\
    AU_1 E &= \frac{\sum_{i=1}^{N} \sum_{x=C_{i}^{AU_1X_{l}}}^{ C_{i}^{AU_1X_{r}}} \sum_{y=C_{i}^{AU_1Y_{t}}}^{ C_{i}^{AU_1Y_{b}}} CA_i(x,y)}{N * S}\\[10pt]
    AU_2 E &= \frac{\sum_{i=1}^{N} \sum_{x=C_{i}^{AU_2X_{l}}}^{ C_{i}^{AU_2X_{r}}} \sum_{y=C_{i}^{AU_2Y_{t}}}^{ C_{i}^{AU_2Y_{b}}} CA_i(x,y)}{N * S}\\
    ...\\
    AU_{43} E &= \frac{\sum_{i=1}^{N} \sum_{x=C_{i}^{AU_{43}X_{l}}}^{ C_{i}^{AU_{43}X_{r}}} \sum_{y=C_{i}^{AU_{43}Y_{t}}}^{ C_{i}^{AU_{43}Y_{b}}} CA_i(x,y)}{N * S}\\
    \end{split}
\end{equation}

% 使用活跃的 AU 的强度作为输入, 对疼痛强度进行回归
\subsection{Pain Intensity Regression Model}
\label{sec: Pain regression}
FACS which defines AUs is a standard to systematically classify the facial expression of emotions\cite{cohn2007observer}. A large number of methods have been developed using AUs for expression recognition and PSPI criteria was proposed to assess adult pain using a linear combination of active AUs in adult pain\cite{cohn2007observer, 8120021}. AU intensity is also an individual-independent feature which possesses stronger generalization ability. To obtain a validated and applicable infants pain assessment method, we use AU intensity for pain intensity regression. To clarify which AUs are active during infant pain, experiments using different number of AUs with reference to their engagement levels are conducted. We finally selected 7 core AUs in infant pain (AU6, AU9, AU10, AU12, AU20, AU25 and AU27) and the related experimental content is presented in Section \ref{sec:AUSelection}. In addition, inspired by the attention mechanism and saliency, engagement levels are also used to facilitate pain intensity assessment. 

As shown in Fig \ref{fig:StructureFig}, the Pain Intensity Regression Model consists of three layers. The model is implemented by multi-layer perceptron (MLP) and intensities of core AUs are used as input. The first layer is the input layer, the size of which is the number of core AUs selected. The second layer is the hidden layer of size 3. The third layer is the output layer which estimates the pain intensity of the infant. The AU intensities and engagement levels are fused by element wise multiplication in input layer. ReLU activation function is used to learn the non-linear relationship between AUs. Dropout is also applied to prevent over-fitting during training.

% 数据集部分
\section{Datasets}
\label{datasets}
% 这部分介绍数据集, 包括其数据分布, 数据介绍, 预处理方法
Three infant pain datasets are used in experiments, the YouTube Immunization dataset, the YouTube Blood Test dataset, and the iCOPEVid (infant Classification Of Pain Expressions Video) dataset.

\begin{figure}[ht]
    \centering
    \includegraphics[width=0.48\textwidth]{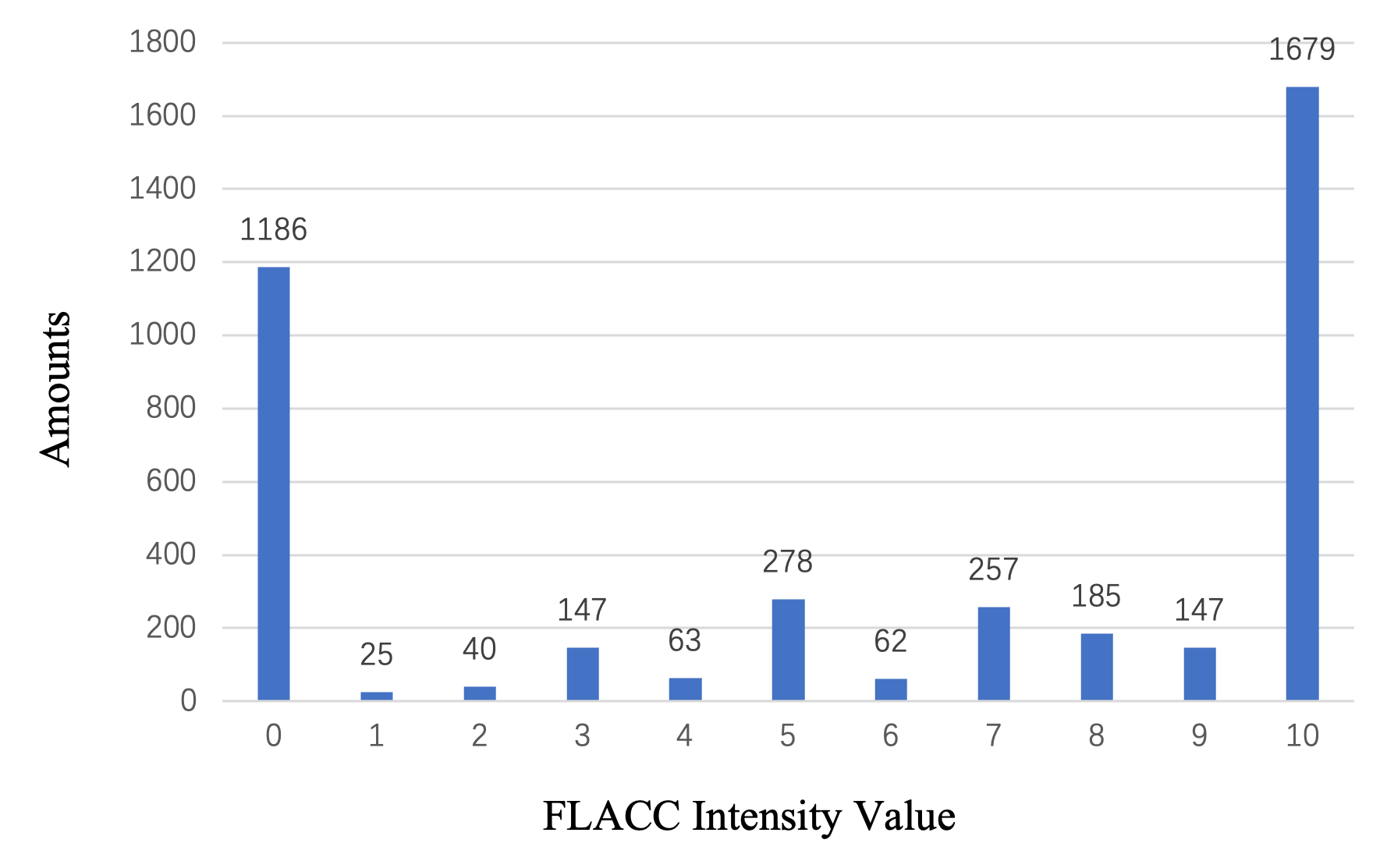}
    \caption{Data distribution of the YouTube Immunization dataset\cite{harrison2014too}. As illustrated in the figure, FLACC scores range from 0-10 and the data set is balanced with the ratings concentrated at 0 or 10.}
    \label{fig:YouTubeImmDis}
\end{figure}

The YouTube Immunization dataset\cite{harrison2014too} aimed to review the content of YouTube videos showing infants being immunized. The researchers wanted to ensure whether pain management measures were taken by medical professionals and parents during the immunization process as well as to observe infants' pain and stress. The review of YouTube videos showing injections in infants less than 12 months was completed using the search terms “baby injection” and “baby vaccine”. Pain was assessed by pain scores using the FLACC (Face, Legs, Activity, Consolability, Cry) tool. FLACC metric ranges from 0 to 10, with pain measures ranging from no pain to very much pain. A total of 142 videos were included and coded by two trained individual viewers. Viewers scored three key time points for each video: 15s before the first injection, at the first injection, at the last injection and 15s after the last injection. Due to the fact that some of the videos are no longer available on YouTube (non-public or offline) and some of the time points in the videos where the baby's face could not be observed, we only collected the videos that are still publicly available and extracted 30 consecutive frames from each observed time point. We finally collected a total of 76 available videos and 4069 flacc scoring frames. We divided the pain into four levels using flacc scores in intervals of 2.5, and the distribution of the data is shown in Table \ref{tab:DatasetDis} upper part and Fig \ref{fig:YouTubeImmDis}.

The YouTube Blood Test dataset\cite{harrison2018systematic} is similar to the YouTube Immunization dataset. It reviewed YouTube videos of babies undergoing blood test to ensure pain management measures were correctly and access pain. A total of 55 videos showing 63 procedures were included and they used Neonatal Facial Coding System (NFCS) scores. As mentioned earlier, NFCS is a system for analyzing infants' facial expressions of pain. NFCS scores ranges from 0 to 4, corresponding to pain intensities ranging from no pain to very painful. Three time points were scored in this dataset: before, during and following the procedure. We likewise collected publicly available videos and extracted 30 consecutive frames from each observed time point at which baby's face could be observed. We finally obtained 19 videos with 396 NFCS scoring frames. Since pain scores of the final collected frames are only 0 or 4, we classified the frames into pain and non-pain levels. The distribution of the dataset is presented in the middle part of Table \ref{tab:DatasetDis}.

The iCOPEVid dataset\cite{brahnam2020neonatal} was to obtain a representative but highly challenging set of video sequences of infant pain facial expressions. This dataset was performed by sequentially changing the infant's crib, blowing air over the face, wiping the infant's extremities with alcohol wipes, and performing a heel puncture for blood collection. Infants were allowed to rest between these procedures. Testing and filming of the babies were done with the consent of the parents and compliance of the ethics protocol of the hospital. The challenge of the dataset is infants have stressful expressions that are not only associated with painful stimuli. Infants will have nervous expressions during many situations which are common in neonatal nursery, including blood testing for heel puncture, alcohol wiping, diaper changes, and crib changes. Therefore, identifying whether facial expression are pain-related is challenging. The dataset contains 49 painful video clips (heel puncture for blood collection, one for each child) and 185 non-painful video clips (several for each child). We preprocessed the video clips and there were finally 445 non-painful frames and 183 painful frames collected. The distribution of the dataset is presented in the third row in Table \ref{tab:DatasetDis} lower part.

% 各数据集分布状况表
\begin{table*}[ht]
\small
\centering
\caption{Data distribution of dataset}
\label{tab:DatasetDis}
\setlength{\tabcolsep}{5pt}{
\renewcommand{\arraystretch}{1.5}
\begin{tabular}{| l | l | l |}
\hline
\textbf{FLACC score in YouTube Immunization Dataset} & \textbf{Pain levels} & \textbf{Number of frames}\\
\hline
$[0, 2.5)$ & No pain & 1251\\
$[2.5, 5)$ & Weak pain & 488\\
$[5, 7.5)$ & Mid pain & 319\\
$[7.5, 10]$ & Strong pain & 2011\\
\hline
\hline
\textbf{NFCS score in YouTube Blood Test Dataset} & \textbf{Pain levels} & \textbf{Number of frames}\\
\hline
0 & No pain & 220\\
1 & - & -\\
2 & - & -\\
3 & - & -\\
4 & Strong pain & 176\\
\hline
\hline
\textbf{Pain score in iCOPEVId dataset} &\textbf{Pain levels} & \textbf{Number of frames}\\
\hline
/ & No pain & 220\\
/ & pain & 176\\
\hline
\end{tabular}}
\end{table*}

% YouTube Immunization Dataset 的分布图

% 实验结果部分
\section{Experimental Results}
% 预处理实验结果的选择
\subsection{Selection of Backbones}
The different backbones are experimented for our end-to-end pain assessment model. Since the YouTube Immunization dataset is derived from real immunization videos uploaded and has the largest amount of data, we use it for the experiments. 5-fold cross-validation method is applied, i.e., we randomly divide the 76 children into 4 folds of number 16 and one fold of number 12. One of the folds is used as the test set each time, and the other four folds are used as the training set. The experimental results are finally averaged over five experiments. This approach is more in line with the clinical setting as it tests whether the model could effectively recognize the facial expressions of different infants.

As shown in Fig \ref{fig:StructureFig}, we pre-process the video frames and use the processed images (infant faces, $ 224 \times 224 \times 3 $) as the input to the pre-trained model. The indicator taken for the experiment are pain levels according to Table \ref{tab:DatasetDis} the upper part; The FLACC values are classified into corresponding pain levels. In training phrase, we use Adam optimizer with initial learning rate of 0.001 and weight decay of 5e-4 to fine-tune pre-trained models. The batch-size is set to 8 and a total of 25 epochs are trained.

The accuracy and F1 score are used as experimental metrics, and the formulas for accuracy and F1 score are as follows. 
\begin{equation} % 分类模型实验所采用的指标 这里需要补充实验指标的说明和意义
    \label{equ:Metric}
    \begin{split}
        &Acc = \frac{TP + TN}{TP + TN + FP + FN} = \frac{TP + TN}{P + N}\\
        &Recall = \frac{TP}{TP + FN}\\
        &Precision = \frac{TP}{TP + FP}\\
        &F1 = \frac{2 \cdot Precision \cdot Recall}{Precision + Recall} = \frac{2TP}{2TP + FN + FP}
    \end{split}
\end{equation}

% 这里还应该补充一些更多的实验结果的说明
The experimental results are shown in Table \ref{tab:PretrainedModeleffect}. The results show that the ResNet18 model has the best performance on the dataset, so we choose this pre-trained model as the backbone in the subsequent experiments.

\subsection{AU Engagement Levels Calculation}
To understand what has been learned by our end-to-end pain assessment model, we intercept the backbone before its fully connected module to compute the Grad-CAM output of the model. According to Section \ref{sec:Engage}, the engagement of different AUs in infant pain then is calculated by using the pixel values of the region around the AU Center in Grad-CAM. As shown in Fig \ref{fig:AUEngagement}, the different engagement levels of AUs are normalized and ordered from biggest to smallest.

It can be observed that during infant pain, the AU associated with mouth movements have a higher correlation compared to adults, such as AU27 AU25. In the PSPI pain formula, the pain is more associated with eye and facial movements such as AU4 AU6 AU7, as expression of adult is more subtle and convergent emotional. In contrast, infants express their feelings directly through facial movements such as grinning or crying. Infants' facial pain expressions are significantly different from those of adults, which is consistent with our hypothesis that infants' pain expressions are more natural and less influenced by social and postnatal development.

% 预训练模型分布表
\begin{table*}[ht]
\small
\centering
\caption{Comparison of different pre-trained models fine-tuned on the YouTube Immunizaition dataset}
\label{tab:PretrainedModeleffect}
\setlength{\tabcolsep}{7pt}{
\renewcommand{\arraystretch}{1.5}
\begin{tabular}{| c | c | c | c | c | c | c | c | c | c |}
\hline
\textbf{Model} & ResNet18 & ResNet34 & ResNet50 & vgg11 & vgg16 & vgg19 & vgg11\_bn & vgg16\_bn & vgg19\_bn\\
\hline
\textbf{Weighted Accuracy(\%)} & \textbf{79.9} & 77.7 & 77.1 & 77.1 & 74.3 & 72.7 & 77.2 & 77.1 & \underline{78.8}\\
\hline
\textbf{Unweighted Accuracy(\%)} & 53.2 & 48.7 & \textbf{57.2} & 54.7 & 47.6 & 46.3 & 49.0 & \underline{55.0} & 48.7\\
\hline
\textbf{Weighted Precision(\%)} & \underline{75.2} & 69.4 & 70.7 & 68.2 & 63.7 & 61.7 & 66.5 & 69.2 & \textbf{78.1}\\
\hline
\textbf{Weighted Recall(\%)} & \textbf{79.9} & 77.7 & 77.1 & 77.1 & 74.3 & 72.7 & 77.2 & 77.1 & \underline{78.8}\\
\hline
\textbf{Weighted F1 score} & \textbf{0.765} & 0.731 & \underline{0.734} & 0.702 & 0.681 & 0.661 & 0.707 & 0.704 & 0.707\\
\hline
\end{tabular}}
\end{table*}

% 不同 AU 的参与程度
\textbf{\begin{figure}[ht]
    \centering
    \includegraphics[width=0.5\textwidth]{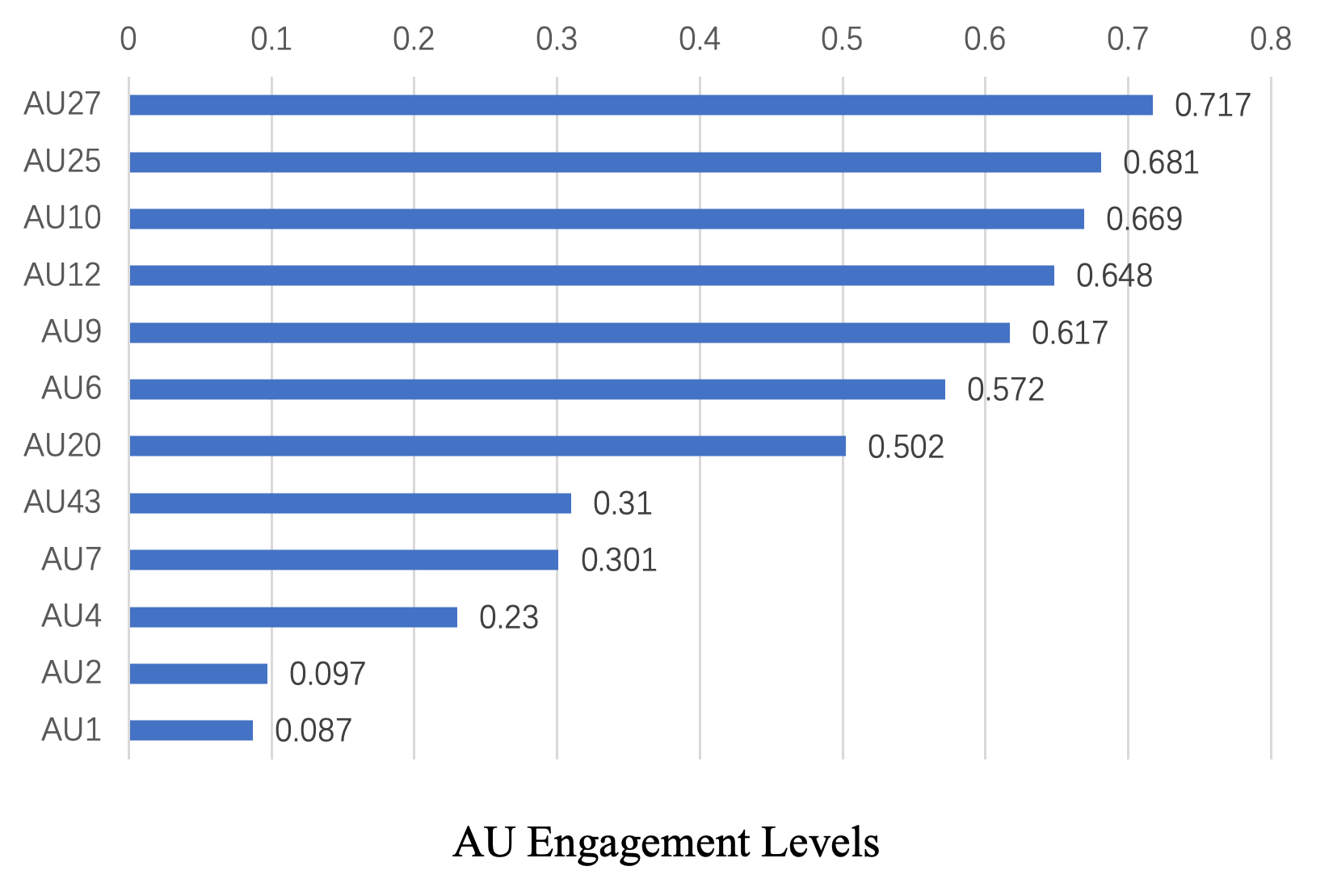}
    \caption{Different engagement levels of AUs in infant pain. Infant's AUs associated with the mouth are more active in pain.}
    \label{fig:AUEngagement}
\end{figure}}

% 对比实验结果
\begin{table}[ht]
\small
\centering
\caption{Comparison of different methods on the YouTube Immunization datasets}
\label{tab:MethodCompareYI}
\setlength{\tabcolsep}{2pt}{
\renewcommand{\arraystretch}{1.5}
\begin{tabular}{| c | c | c | c | c | c |}
\hline
\textbf{Method} & \textbf{WA(\%)} & \textbf{UA(\%)} & \textbf{Precision(\%)} & \textbf{Recall(\%)} & \textbf{F1 Score}\\
\hline
PSPI & 68.7 & 67.9 & 68.7 & 68.7 & 0.684\\
ResNet18 & 79.9 & 53.2 & 75.2 & 79.9 & 0.765\\
ResNet34 & 77.7 & 48.7 & 69.4 & 77.7 & 0.731\\
vgg11bn & 77.2 & 49.0 & 66.5 & 77.2 & 0.707\\
vgg16bn & 77.1 & 55.0 & 69.2 & 77.1 & 0.704\\
vgg19bn & 78.8 & 48.7 & 78.1 & 78.8 & 0.707\\
12 AU & 81.9 & 83.1 & 83.8 & 81.9 & 0.821\\
7 AU & \underline{85.3} & \underline{85.5} & \underline{86.2} & \underline{85.3} & \underline{0.853}\\
7 AU with weight & \textbf{90.3} & \textbf{90.3} & \textbf{91.2} & \textbf{90.3} & \textbf{0.902}\\
\hline
\end{tabular}}
\end{table}

\subsection{Pain Intensity Regression}
% 简单描述实验原因
The pain intensity of adult in pictures can be quickly assessed by PSPI. To obtain a validated pain assessment method similar to PSPI but applicable to infants, we use AU intensity for pain intensity regression. Specific details of the regression model can be found in Section \ref{sec: Pain regression}. The regression model is trained on the YouTube Immunization dataset, using the same data distribution as the above experiment. We use Adam optimizer with initial learning rate of 0.01 and SmoothL1Loss loss function. The batch-size is set to 8 and a total of 100 epochs are trained.

% 筛选 AU 的分布图
\begin{figure}[ht]
    \centering
    \includegraphics[width=0.48\textwidth]{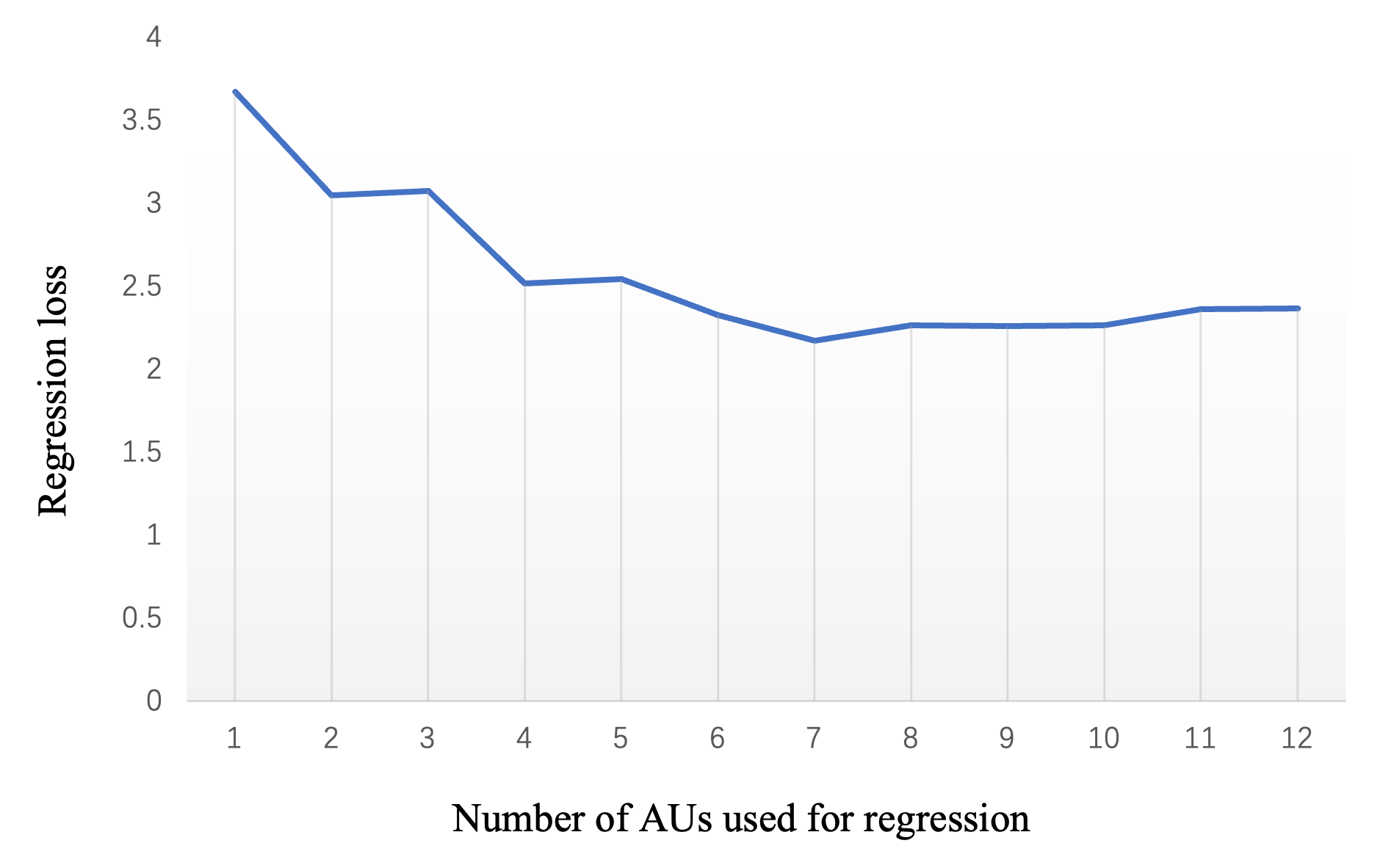}
    \caption{Experimental comparison of regression using different numbers of AUs according to engagement ranking. When top 7 AUs (AU27, AU 25, AU 10, AU 12, AU 9, AU6, AU20) are used for pain intensity regression, the loss is the lowest.}
    \label{fig:AUSelection}
\end{figure}

% AU 选择的实验
\subsubsection{Selection of core AUs}
\label{sec:AUSelection}
In previous practices of using FACS for pain assessment, they only select specific AUs based on previous experience and neglect different engagement levels of AUs in pain expression. In contrast, core AUs which are closely associated with pain expressions are selected according to their engagement ranking.

The ranking of AU engagement levels is displayed in Fig \ref{fig:AUEngagement}. A different numbers of AUs are used for pain intensity regression. As shown in the Fig \ref{fig:AUSelection}, experimental results show that when these AUs (AU27, AU 25, AU 10, AU 12, AU 9, AU6, AU20) are used, the lowest loss can be achieved on the test set. Therefore, we selected these AUs as the core AUs related to infant pain.

% 不同方法的实验
\subsubsection{Method comparison}
Experimental results in Table \ref{tab:MethodCompareYI} illustrates performance of different methods on YouTube Immunization dataset. The PSPI metric performance is inferior to the end-to-end models and the AU pain intensity regression model. This is because the PSPI model is derived from the facial expressions of pain in adults, which is significantly different from those of infants. Infants express pain more directly and intensely. 

The experimental results also illustrate that the regression model has better performance than the end-to-end models. In particular, the UA results in Table \ref{tab:MethodCompareYI} show that the AU model also performs well with unbalanced datasets. The core AU associated with infant pain has also been shown to have better performance compared to using all AUs. When we use engagement levels to facilitate pain intensity assessment (Section \ref{sec: Pain regression}), the model achieves the best results. The above experimental results prove their close association between AUs and facial pain expressions, and the importance of AU engagement in infant pain.

% 进一步的验证
\subsection{Validation of generalization ability}
Both our neural network model and regression model are only trained on the YouTube Immunization dataset. For further validation of generalization ability in a more realistic clinical setting, two unseen datasets (YouTube Blood Test and iCOPEVid datasets\cite{harrison2018systematic,brahnam2020neonatal}) are tested on our trained models without any fine-tuning or further training.
\begin{table}[ht]
\small
\centering
\caption{Comparison of different methods on the iCOPEVid datasets}
\label{tab:MethodCompareCOPE}
\setlength{\tabcolsep}{2pt}{
\renewcommand{\arraystretch}{1.5}
\begin{tabular}{| c | c | c | c | c | c |}
\hline
\textbf{Method} & \textbf{WA(\%)} & \textbf{UA(\%)} & \textbf{Precision(\%)} & \textbf{Recall(\%)} & \textbf{F1 Score}\\
\hline
PSPI & 48.9 & 56.3 & 65.4 & 48.9 & 0.501\\
ResNet18 & \underline{65.7} & \textbf{72.3} & \underline{78.5} & \underline{65.7} & 0.671\\
ResNet34 & 62.9 & \textbf{72.3} & \textbf{80.7} & 62.9 & 0.640\\
vgg11bn & \underline{65.7} & \underline{71.8} & 77.8 & \underline{65.7} & \underline{0.672}\\
vgg16bn & 63.9 & 68.9 & 75.1 & 63.9 & 0.655\\
vgg19bn & 59.1 & 65.8 & 73.5 & 59.1 & 0.606\\
% 12AU & 62.8 & 68.6 & 75.2 & 62.8 & 0.644\\
% 7AU & 70.4 & 73.3 & 77.7 & 70.4 & 0.718\\
AuE-IPA & \textbf{67.4} & \textbf{72.3} & 77.7 & \textbf{67.4} & \textbf{0.689}\\
\hline
\end{tabular}}
\end{table}

\begin{table}[ht]
\small
\centering
\caption{Comparison of different methods on the YouTube Blood Test datasets}
\label{tab:MethodCompareYB}
\setlength{\tabcolsep}{2pt}{
\renewcommand{\arraystretch}{1.5}
\begin{tabular}{| c | c | c | c | c | c |}
\hline
\textbf{Method} & \textbf{WA(\%)} & \textbf{UA(\%)} & \textbf{Precision(\%)} & \textbf{Recall(\%)} & \textbf{F1 Score}\\
\hline
PSPI & 68.3 & 64.2 & 70.1 & 68.3 & 0.655\\
ResNet18 & 67.0 & 70.4 & 76.0 & 67.0 & 0.660\\
ResNet34 & 72.8 & 71.5 & 72.6 & 72.8 & 0.726\\
vgg11bn & 71.3 & \underline{74.6} & \underline{80.0} & 71.3 & 0.706\\
vgg16bn & \underline{74.4} & 72.5 & 74.4 & \underline{74.4} & \underline{0.739}\\
vgg19bn & 69.5 & 73.6 & \textbf{82.3} & 69.5 & 0.681\\
% 12AU & 75.6 & 77.2 & 78.6 & 75.6 & 0.757\\
% 7AU & 75.6 & 78.7 & 83.8 & 75.6 & 0.752\\
AuE-IPA & \textbf{76.4} & \textbf{75.6} & 76.3 & \textbf{76.4} & \textbf{0.763}\\
\hline
\end{tabular}}
\end{table}

Since the three datasets do not use a uniform pain annotation, pain levels corresponding to the datasets are used for comparison as shown in Table \ref{tab:DatasetDis}. The same evaluation metrics as the above experiments are used in validation.

In the comparison between the neural network model and the AU pain intensity regression model in Table \ref{tab:MethodCompareYB} and Table \ref{tab:MethodCompareCOPE}, it can be seen that the facial model lags behind the regression model for unfamiliar data due to poor generalization ability. Since the regression model takes AU intensity as input, and AU intensity is an individual-independent feature that possesses stronger generalization ability. Therefore, the regression model works better in wild and unfamiliar scenarios, as evidenced by experimental results.

% 实验结论 欠缺一些对未来工作的展望 现有工作的不足
\section{Conclusion}

Automatic assessment and proper management of infant pain can effectively reduce the psychological and brain damage caused by pain. In this paper, after analyzing the deficiencies of the popular pain detection standard PSPI and current infant pain assessment approaches, we propose a novel AuE-IPA method is proposed. Based on Grad-CAM output of neural network models, the different engagement levels of facial AUs in infant pain are revealed. According to engagement ranking, the 7 core AUs are selected. A new regression model using core AUs and their engagement is proposed to assess infant pain. Experimental results on the YouTube dataset illustrate that our pain intensity regression method performs better compared to PSPI and end-to-end deep networks. In the further validation on two unseen datasets, experiments show that our method possesses a stronger generalization ability.

\section*{Acknowledgment}
% 这里暂时用于测试引用

\ifCLASSOPTIONcaptionsoff
  \newpage
\fi

\bibliographystyle{IEEEtran}
\bibliography{refs}

\end{document}